# The RSNA Abdominal Traumatic Injury CT (RATIC) Dataset

**Authors:** Jeffrey D. Rudie, Hui-Ming Lin, Robyn L. Ball, Sabeena Jalal, Luciano M. Prevedello, Savvas Nicolaou, Brett S. Marinelli, Adam E. Flanders, Kirti Magudia, George Shih, Melissa A. Davis, John Mongan, Peter D. Chang, Ferco H. Berger, Sebastiaan Hermans, Meng Law, Tyler Richards, Jan-Peter Grunz, Andreas Steven Kunz, Shobhit Mathur, Sandro Galea-Soler, Andrew D. Chung, Saif Afat, Chin-Chi Kuo, Layal Aweidah, Ana Villanueva Campos, Arjuna Somasundaram, Felipe Antonio Sanchez Tijmes, Attaporn Jantarangkoon, Leonardo Kayat Bittencourt, Michael Brassil, Ayoub El Hajjami, Hakan Dogan, Muris Becircic, Agrahara G. Bharatkumar, Eduardo Moreno Júdice de Mattos Farina, **Dataset Curator Group**, **Dataset Contributor Group**, **Dataset Annotator Group**, Errol Colak.

**Affiliations:** Department of Radiology, Scripps Clinic Medical Group and University of California San Diego (J.D.R.), Department of Medical Imaging, St. Michael's Hospital, Unity Health Toronto (H.M.L., S.H., S.M., E.C.), The Jackson Laboratory, Bar Harbor, ME (R.L.B.), Department of Radiology, Vancouver General Hospital, Vancouver, Canada, Department of Radiology, (S.J., S.N.), Department of Radiology, The Ohio State University, Columbus, Ohio (L.M.P.), Memorial Sloan Kettering Cancer Center, New York, NY (B.S.M.), Department of Radiology, Thomas Jefferson University, Philadelphia, PA (A.E.F.), Duke University School of Medicine, Durham, NC (K.M.), Department of Radiology, Weill Cornell Medicine, New York, NY (G.S.), Department of Radiology and Biomedical Imaging, Yale University School of Medicine, New Haven, CT (M.A.D.), Department of Radiology and Biomedical Imaging, University of California San Francisco, San Francisco, CA (J.M.), Departments of Radiological Sciences and Computer Science, University of California, Irvine, CA (P.D.C.), Sunnybrook Health Sciences Centre, University of Toronto (F.H.B), Alfred Health, Monash University, Melbourne, Australia (M.L.), University of Utah, Salt

Lake City, Utah (T.R.), Department of Diagnostic and Interventional Radiology, University Hospital of Würzburg, Würzburg, Germany (JP.G., A.S.K.), Medical Imaging Department, Mater Dei Hospital, Msida, Malta (S. GS.), Department of Diagnostic Radiology, Queen's University, Kingston, Canada (A.D.C.), School of Computing, Queen's University (M.H.), Department of Diagnostic and Interventional Radiology, Eberhard-Karls-University Tübingen, Tübingen, Germany (S.A.), Division of Nephrology, Department of Internal Medicine, China Medical University Hospital and College of Medicine, China Medical University, Taichung, Taiwan (C.C.K.), Big Data Center, China Medical University Hospital, Taichung, Taiwan (C.C.K.), AKI-CARE (Clinical Advancement, Research and Education) Center, Department of Internal Medicine, China Medical University Hospital, Taichung, Taiwan (C.C.K.), Department of Medical Imaging Liverpool Hospital, Sydney, Australia (L.A.), Hospital Universitario Ramón y Cajal, Madrid, Spain (A.V.C.), Department of Radiology, Gold Coast University Hospital, Griffith University, Gold Coast, Queensland, Australia (A.S.), Department of Medical Imaging, Clínica Santa María, Santiago, Chile (F.A.S.T.), Department of Radiology, Faculty of Medicine, Chiang Mai University, Chiang Mai, Thailand (A.J.), Department of Radiology, University Hospitals Cleveland Medical Center and Associate Professor, Case Western Reserve University School of Medicine, Cleveland, OH (L.K.B.), Department of Radiology, Tallaght University Hospital, Dublin, Ireland (M.B.), Radiology Department, Arrazi Hospital, CHU Mohamed VI Cadi Ayyad University, Marrakech, Morocco (A.E.H.), Department of Radiology, Koç University School of Medicine, Istanbul, Türkiye (H.D.), Clinical Center University of Sarajevo, Sarajevo, Bosnia and Herzegovina (M.B.), Department of Diagnostic Radiology, Medical College of Wisconsin, Milwaukee, WI (A.G.B.), Department of Diagnostic Imaging, Universidade Federal de São Paulo, São Paulo, Brazil (E.M.J.M.F.), Department of Medical Imaging, University of Toronto, Toronto, Canada  (S.M., E.C.),

***Dataset Curator Group:***

Matthew Aitken, Patrick Chun-Yin Lai, Priscila Crivellaro, Jayashree Kalpathy-Cramer, Zixuan Hu, Reem Mimish, Aeman Muneeb, Mitra Naseri, Maryam Vazirabad, Rachit Saluja

***Dataset Contributor Group:***

Nitamar Abdala, Jason Adleberg, Waqas Ahmad, Christopher O. Ajala, Emre Altinmakas, Robin Ausman, Miguel Ángel Gómez Bermejo, Deniz Bulja, Jeddi Chaimaa, Lin-Hung Chen, Sheng-Hsuan Chen, Hsiu-Yin Chiang, Rahin Chowdhury, David Dreizin, Zahi Fayad, Yigal Frank, Sirui Jiang, Belma Kadic, Helen Kavnoudias, Alexander Kagen, Felipe C. Kitamura, Nedim Kruscica, Michael Kushdilian, Brian Lee, Jennifer Lee, Robin Lee, Che-Chen Lin, Karun Motupally, Eamonn Navin, Andrew S. Nencka, Christopher Newman, Akdi Khaoula, Shady Osman, William Parker, Jacob J. Peoples, Marco Pereañez, Christopher Rushton, Navee Sidi Mahmoud, Xueyan Mei, Beverly Rosipko, Muhammad Danish Sarfarz, Adnan Sheikh, Maryam Shekarforoush, Amber Simpson, Ashlesha Udare, Victoria Uram, Emily V. Ward, Conor Waters, Min-Yen Wu, Wanat Wudhikulprapan, Adil Zia

***Dataset Annotator Group:***

Claire K. Sandstrom, Angel Ramon Sosa Fleitas, Joel Kosowan, Christopher J Welman, Sevtap Arslan, Mark Bernstein, Linda C. Chu, Karen S. Lee, Chinmay Kulkarni, Taejin Min, Ludo Beenen, Betsy Jacobs, Scott Steenburg, Sree Harsha Tirumani, Eric Wallace, Shabnam Fidvi, Helen Oliver, Casey Rhodes, Paulo Alberto Flejder, Adnan Sheikh, Muhammad Munshi, Jonathan Revels, Vinu Mathew, Marcela De La Hoz Polo, Apurva Bonde, Ali Babaei Jandaghi, Robert Moreland, M. Zak Rajput, James T. Lee, Nikhil Madhuripan, Ahmed Sobieh, Bruno Nagel Calado, Jeffrey D Jaskolka, Lee Myers, Laura Kohl, Matthew Wu, Wesley Chan, Facundo Nahuel Diaz

**Abstract:**

The RSNA Abdominal Traumatic Injury CT (RATIC) dataset is the largest publicly available collection of adult abdominal CT studies annotated for traumatic injuries. This dataset includes 4,274 studies from 23 institutions across 14 countries. The dataset is freely available for non-commercial use via Kaggle at https://www.kaggle.com/competitions/rsna-2023-abdominal-trauma-detection. Created for the RSNA 2023 Abdominal Trauma Detection competition, the dataset encourages the development of advanced machine learning models for detecting abdominal injuries on CT scans. The dataset encompasses detection and classification of traumatic injuries across multiple organs, including the liver, spleen, kidneys, bowel, and mesentery. Annotations were created by expert radiologists from the American Society of Emergency Radiology (ASER) and Society of Abdominal Radiology (SAR). The dataset is annotated at multiple levels, including the presence of injuries in three solid organs with injury grading, image-level annotations for active extravasations and bowel injury, and voxelwise segmentations of each of the potentially injured organs. With the release of this dataset, we hope to facilitate research and development in machine learning and abdominal trauma that can lead to improved patient care and outcomes.

**Introduction**

Trauma is the most common cause of fatal injuries in Americans under the age of 45 and claims six million lives globally each year (1). Early accurate diagnosis and grading of traumatic injuries is critical in guiding clinical management and improving patient outcomes. Computed tomography plays a central role in the initial evaluation of hemodynamically stable patients (2,3). For blunt and penetrating abdominal trauma, the American Association for the Surgery of Trauma (AAST) organ injury grading system is the most well recognized system for grading solid organ injuries (4,5) and is critical in triaging patients between surgery, minimally invasive intervention, and conservative management (6).

While the AAST grading system is an important guide to assess solid organ injury, rapid interpretation of trauma studies is challenging given the large number of images needed to review and potential for subtle findings. In fact, diagnostic errors in the interpretation of trauma are common (7) and there is high inter-rater variability in the AAST grading system (8,9). Furthermore, the large variation in protocols used at different hospitals, including a single portal venous phase, multiphasic imaging, and split bolus approaches (10), can further complicate this task.

Automated assessment of traumatic abdominal injuries is an excellent use case for artificial intelligence (AI) algorithms given the potential to prioritize studies that may require more expedient interpretations as well as augment radiologist accuracy and efficiency, which may be particularly true in areas where subspecialists are in short supply. Recent work on AI based assessment of abdominal trauma includes studies on automated detection of splenic (11-14) and liver (15) injury, hemoperitoneum (16), and pneumoperitoneum (17). However, prior studies have typically been limited in scope to single organs and single institutions, and realistically not generalizable into clinical practice. Thus, there is a need for large multi-institutional publicly available annotated abdominal trauma datasets to address this challenge.

The Radiological Society of North America (RSNA) collaborated with the American Society of Emergency Radiology (ASER) and Society of Abdominal Radiology (SAR) to curate a large, publicly available expert-labeled dataset of abdominal CT images for traumatic injuries focusing on injuries to the liver, spleen, kidneys, bowel and mesentery, and active extravasation. This dataset was used for the RSNA 2023 Abdominal Trauma Detection competition which attracted 1,500 competitors from around the world to develop innovative machine learning (ML) models that detect traumatic injuries on abdominal CT.

**Dataset Description and Usage**

The RSNA Abdominal Traumatic Injury CT (RATIC) dataset is composed of CT scans of the abdomen and pelvis from 4,274 patients with a total of 6,481 image series from 23 institutions across 14 countries and 6 continents. A detailed breakdown of patient demographics and injuries across the different institutions are provided in **Table 1**. The composition of the demographics and injuries are found in **Table 2** with a breakdown of injury severity in **Table 3**.

CT images are in Digital Imaging and Communications in Medicine (DICOM) format. Study level injury annotations and demographic information are provided in four comma-separated value files. The train_2024.csv file contains information about the presence of traumatic abdominal injuries (liver, kidney, spleen, bowel/mesenteric and active extravasation) for each patient. The image_level_labels_2024.csv file provides image level labels for bowel/mesenteric injuries and active extravasation. The train_series_meta_2024.csv file contains information regarding the phase of imaging and anatomical coverage of each CT series. The train_demographics_2024.csv file contains information about patient demographics. Pixel-level segmentations of abdominal organs are provided in Neuroimaging Informatics Technology Initiative (NIfTI) format for a subset of 206 series from the training set.

A flowchart of how the RATIC dataset was curated and annotated is shown in **Figure 1** with a detailed description provided in **Supplementary Materials**. In brief, sites provided initial labels for the presence of different traumatic injuries based on clinical reports. Radiologist annotators recruited from the ASER and SAR then annotated solid organ injury grades and locations of bowel/mesenteric injuries and active extravasation. Ground truth labels for the grading of solid organ injuries were established through best of 3 majority voting and collapsed into low (AAST I - III) and high grade (IV and V) injury groups. Image level labels for bowel/mesenteric injuries and active extravasation were based on the consensus of different annotators. Voxel-wise segmentations (**Figure 2**) were manually corrected after training a nnU-Net (18) on the TotalSegmentator dataset (19), focusing only on the organs being evaluated in the challenge: (1) liver, (2) spleen, (3) left kidney, (4) right kidney, and (5) bowel (representing a combination of esophagus, stomach, duodenum, small bowel, and colon).

**Discussion**

We curated a large, high-quality dataset of abdominal trauma CTs with contributions from 23 institutions in 14 countries and 6 continents. This represents the largest and most diverse publicly available dataset of abdominal trauma CT scans. This dataset provides annotations relating to injuries of the liver, spleen, kidneys, bowel, and mesentery, as well as active extravasation. This rich dataset has further utility for investigators as other injuries such as hematomas, fractures, and lower thoracic injuries are present within the dataset but were not explicitly annotated due to intent of the challenge to focus on critical findings of the highest clinical importance for trauma patients.

We chose broad inclusion criteria for the contributed CT scans. Our initial survey of potential contributing sites showed great variety in the protocols used for imaging abdominal trauma patients. In fact, some institutions had multiple protocols and selected a protocol based on the severity of the trauma. Aspects that varied across protocols included the parts of the body that were imaged, phases of imaging, and slice thickness. Stringent inclusion criteria that limited contributed scans to a single homogenous protocol (e.g. thin slice, multiphasic CT scans of the abdomen and pelvis) would severely constrain the size and potential generalizability of this dataset. For this reason, we widened the inclusion criteria to facilitate a larger and more diverse dataset that could then be used to train more robust ML models. Biphasic (arterial and portal venous), split bolus, and portal venous phase protocols were considered acceptable.

Participating sites were asked to enrich the dataset with representative injuries given the relatively low prevalence of traumatic abdominal injuries on CT encountered in clinical practice. Despite this request, the number of cases with injuries submitted was lower than the organizing committee had anticipated. Addressing class imbalances in curated datasets is particularly important in improving ML model robustness and reducing bias (20,21). An explicit effort was made by the organizing committee to reduce potential biases in the dataset by considering factors such as sex, age, injuries, and contributing site when assigning scans to the train, public test, and private test datasets.

A challenge we faced in curating this dataset was the dramatic differences in z-axis coverage in the contributed CT scans. For example, some sites imaged from the skull vertex to feet while many sites limited imaging to the abdomen and pelvis. To reduce the size of the dataset and to help ML model training by reducing the search space, we decided to limit scans to the abdomen and pelvis, using an upper bound of the mid-heart and lower bound of the proximal femurs using an automated pipeline (22) and manually reviewed the processed scans.

Similar to prior challenges, we wanted to maximize use of the data and ensure high quality labels while not overburdening annotators. Contributing sites pre-labelled submitted scans with information extracted from the clinical report that allowed annotators to focus on the abnormal scans. We considered a variety of annotation strategies that ranged from study to pixel level annotations. Our past experience with the cervical spine fracture detection challenge (23) showed that the prize-winning models relied on study level annotations and segmentations rather than bounding boxes (24). In addition, recent work has shown that strongly supervised models trained on slice level labels from the RSNA Brain CT Hemorrhage dataset labels (25) do not outperform weakly-supervised models trained on study level labels (26). Pixel level annotations of injuries, including bounding boxes, would be a time consuming task with likely poor reproducibility as abdominal injuries can be quite complex with ill-defined borders. We settled on providing segmentations of the relevant abdominal organ systems to assist with localization and organ labels at the study level. Image level labels were provided for bowel/mesenteric injuries and active extravasation as these injuries can be subtle, present in variable anatomic locations, and present on a limited number of images.

Individual annotators were assigned a single organ system to annotate rather than providing annotations for multiple organ systems on their assigned CT scans. The organizing committee felt this would improve the efficiency of the annotation process and label quality by allowing an annotator to focus on a single task and AAST injury grading scale. The annotators provided granular labels using the AAST grading scale for solid organ injuries. Due to the well documented issues with inter-rater agreement in the grading of solid organ injuries with AAST (9) and to help model training, AAST grades I - III were collapsed into low grade injuries while grades IV and V were considered high grade injuries. This grouping of injury grades still provides more information than a binary label for injury and reflects many clinical practices where grade IV and V injuries are more likely to undergo surgery or endovascular treatment (4-6). Rather than assigning a fixed number of CT scans for annotation, we utilized the crowd sourcing mode on the annotation platform. This allowed annotators to label as many cases as they wanted with the public scoreboard providing motivation.

Each solid organ injury label was annotated independently by 3 radiologists and the final ground truth labels were established by majority. In scenarios where all 3 annotators assigned different gradings (i.e. no injury, low grade, and high grade), a member of the organizing committee adjudicated the case and assigned the final ground truth label. We felt that this approach would improve annotation quality by generating labels with better inter-rater agreement and avoiding the problem of a poor-quality annotation in a single annotation scheme. With an approach that relies on a single annotator per scan, it can be difficult to detect poor quality annotators following completion of the trial of training cases.

There are several limitations of this dataset. Ground truth labels for the grading of solid organ injuries were established through best of 3 majority voting. While this represents an improvement over a single annotator, there are inherent issues with AAST grading as a result of inter-rater variability. We recognize the absence of delayed phase imaging as a limitation since it forms part of the AAST imaging criteria for grade II - IV renal injuries in terms of collecting system injuries. Delayed phase imaging was not included as it was not part of the routine protocol for most contributing sites and we were concerned that its inclusion in cases with renal injuries would bias models potentially through spurious associations rather than truly detecting collecting system injuries. Finally, ground truth labels for solid organ injuries, bowel and mesenteric injuries, and active extravasation were made using a web-based annotation platform which is limited when compared to real world clinical practice with access to high resolution monitors, multi-planar thin slice imaging, clinical information, and prior imaging examinations.

In summary, the RATIC dataset represents the largest and most geographically diverse, publicly available expert-annotated dataset of abdominal traumatic injury CT studies. With the release of this dataset, we hope to facilitate research and development in machine learning and abdominal trauma that can lead to improved patient care and outcomes. This dataset is made freely available to all researchers for non-commercial use.

## Tables

| Site ID | Sex | | | Age (y) | Total Cases | Negative Injury | Positive Injury | | | | | |
| | Male | Female | UNK | | | | Total Positive | Liver Injury | Spleen Injury | Kidney Injury | Bowel Injury | Active Extravasation |
|---|---|---|---|---|---|---|---|---|---|---|---|---|
| Site 1 | 195 | 55 | 0 | 52.9±20.6 (18-90) | 250 | 202 | 48 | 12 | 24 | 6 | 9 | 20 |
| Site 2 | 49 | 1 | 45 | 37.1±13.8 (19-71) 37 UNK | 95 | 76 | 19 | 10 | 9 | 4 | 0 | 1 |
| Site 3 | 72 | 22 | 0 | 45.1±16.5 (18-88) | 94 | 76 | 18 | 9 | 8 | 6 | 1 | 2 |
| Site 4 | 200 | 93 | 0 | 41.8±17.8 (20-90) | 293 | 147 | 146 | 48 | 54 | 78 | 17 | 28 |
| Site 5 | 16 | 4 | 0 | 41.8±20.9 (20-85) | 20 | 12 | 8 | 3 | 5 | 5 | 0 | 1 |
| Site 6 | 343 | 190 | 0 | 56.3±21.4 (18-90) | 533 | 436 | 97 | 40 | 37 | 21 | 13 | 38 |
| Site 7 | 141 | 50 | 0 | 47.2±20.0 (18-90) | 191 | 155 | 36 | 11 | 11 | 9 | 7 | 8 |
| Site 8 | 148 | 45 | 0 | 46.2±20.4 (20-90) | 193 | 107 | 86 | 26 | 48 | 23 | 12 | 20 |
| Site 9 | 109 | 51 | 0 | 44.5±19.6 (19-90) | 160 | 119 | 41 | 18 | 12 | 14 | 1 | 8 |
| Site 10 | 96 | 38 | 0 | 43.7±17.8 (18-90) | 134 | 68 | 66 | 29 | 14 | 17 | 8 | 17 |
| Site 11 | 17 | 21 | 0 | 51.9±22.2 (20-90) | 38 | 38 | 0 | 0 | 0 | 0 | 0 | 0 |
| Site 12 | 102 | 31 | 0 | 42.1±17.9 (18-90) | 133 | 85 | 48 | 28 | 15 | 16 | 4 | 6 |
| Site 13 | 127 | 63 | 0 | 49.8±21.6 (18-90) | 190 | 113 | 77 | 33 | 37 | 11 | 6 | 18 |
| Site 14 | 179 | 68 | 0 | 46.9±20.8 | 247 | 135 | 112 | 46 | 52 | 33 | 14 | 43 |

|  |  |  |  | (18-90) |  |  |  |  |  |  |  |  |
|---|---|---|---|---|---|---|---|---|---|---|---|---|
| **Site 15** | 110 | 52 | 0 | 46.2±19.2 (18-90) | 162 | 101 | 61 | 24 | 24 | 13 | 3 | 13 |
| **Site 16** | 117 | 77 | 0 | 46.4±19.7 (18-90) | 194 | 125 | 69 | 17 | 30 | 23 | 2 | 16 |
| **Site 17** | 99 | 34 | 0 | 38.4±16.8 (18-90) | 133 | 61 | 72 | 38 | 26 | 24 | 11 | 17 |
| **Site 18** | 63 | 11 | 0 | 47.9±18.8 (18-85) | 74 | 50 | 24 | 9 | 5 | 5 | 2 | 7 |
| **Site 19** | 116 | 61 | 0 | 56.1±21.1 (18-90) | 177 | 104 | 73 | 13 | 21 | 24 | 3 | 32 |
| **Site 20** | 162 | 145 | 0 | 57.1±20.6 (19-90) 14 UNK | 307 | 232 | 75 | 18 | 27 | 9 | 1 | 20 |
| **Site 21** | 253 | 95 | 0 | 43.5±19.9 (18-90) 3 UNK | 348 | 331 | 17 | 4 | 6 | 5 | 2 | 2 |
| **Site 22** | 68 | 37 | 0 | 44.4±20.6 (18-90) | 105 | 90 | 15 | 8 | 5 | 4 | 0 | 1 |
| **Site 23** | 155 | 48 | 0 | 42.4±18.2 (18-88) | 203 | 95 | 108 | 47 | 47 | 20 | 17 | 18 |
| **Total** | 2,937 | 1,292 | 45 | 48.0±20.6 (18-90) 54 UNK | 4,274 | 2,958 | 1,316 | 491 | 517 | 370 | 133 | 336 |

**Table 1.** The distribution of positive and negative cases for abdominal injury with breakdown of injury class per each institution. Age is represented by mean ± standard deviation and the range is provided in parentheses. The number of cases with unknown age and sex is denoted by UNK.

| Usage | Sex | | | Age (y) | Total Cases | Negative Injury | Positive Injury | | | | | |
|---|---|---|---|---|---|---|---|---|---|---|---|---|
| | Male | Female | UNK | | | | Total Positive | Liver Injury | Spleen Injury | Kidney Injury | Bowel Injury | Active Extravasation |
| Train | 2,164 | 949 | 34 | 47.9±21.0 (18-90) 40 UNK | 3,147 | 2,237 | 910 | 340 | 372 | 217 | 71 | 215 |
| Public | 282 | 121 | 1 | 48.2±19.6 (18-90) 6 UNK | 404 | - | - | - | - | - | - | - |
| Private | 491 | 222 | 10 | 48.0±19.5 (18-90) 8 UNK | 723 | - | - | - | - | - | - | - |

**Table 2.** The distribution of positive and negative cases for abdominal injury with breakdown of injury class for the training portion of the dataset. The composition of the public and private test sets is confidential. The demographic distribution is also provided. Age is represented by mean ± standard deviation and the range is provided in parentheses. The number of cases with unknown age and sex is denoted by UNK. The

| Usage | Liver | | Spleen | | Kidneys | | Bowel Injury Images | Active Extravasation Images |
|---|---|---|---|---|---|---|---|---|
| | Low | High | Low | High | Low | High | | |
| Train | 273 | 67 | 210 | 162 | 141 | 76 | 7,232 | 8,400 |

**Table 3.** Distribution of injuries for the training portion of the dataset.

**Figures**

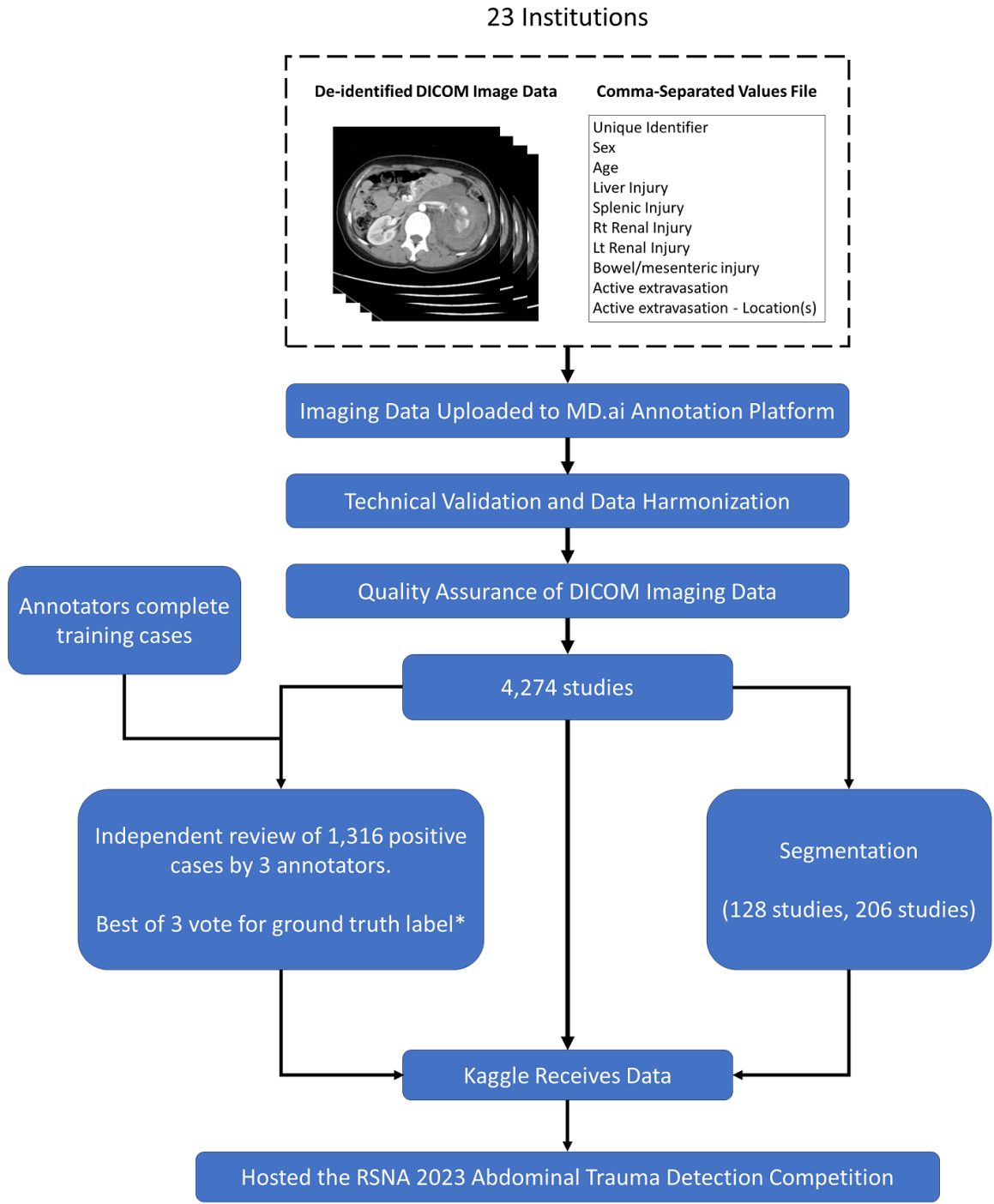

**Figure 1**: Summary of the data curation and annotation process. * Bowel/mesenteric injuries were reviewed by 2 annotators.

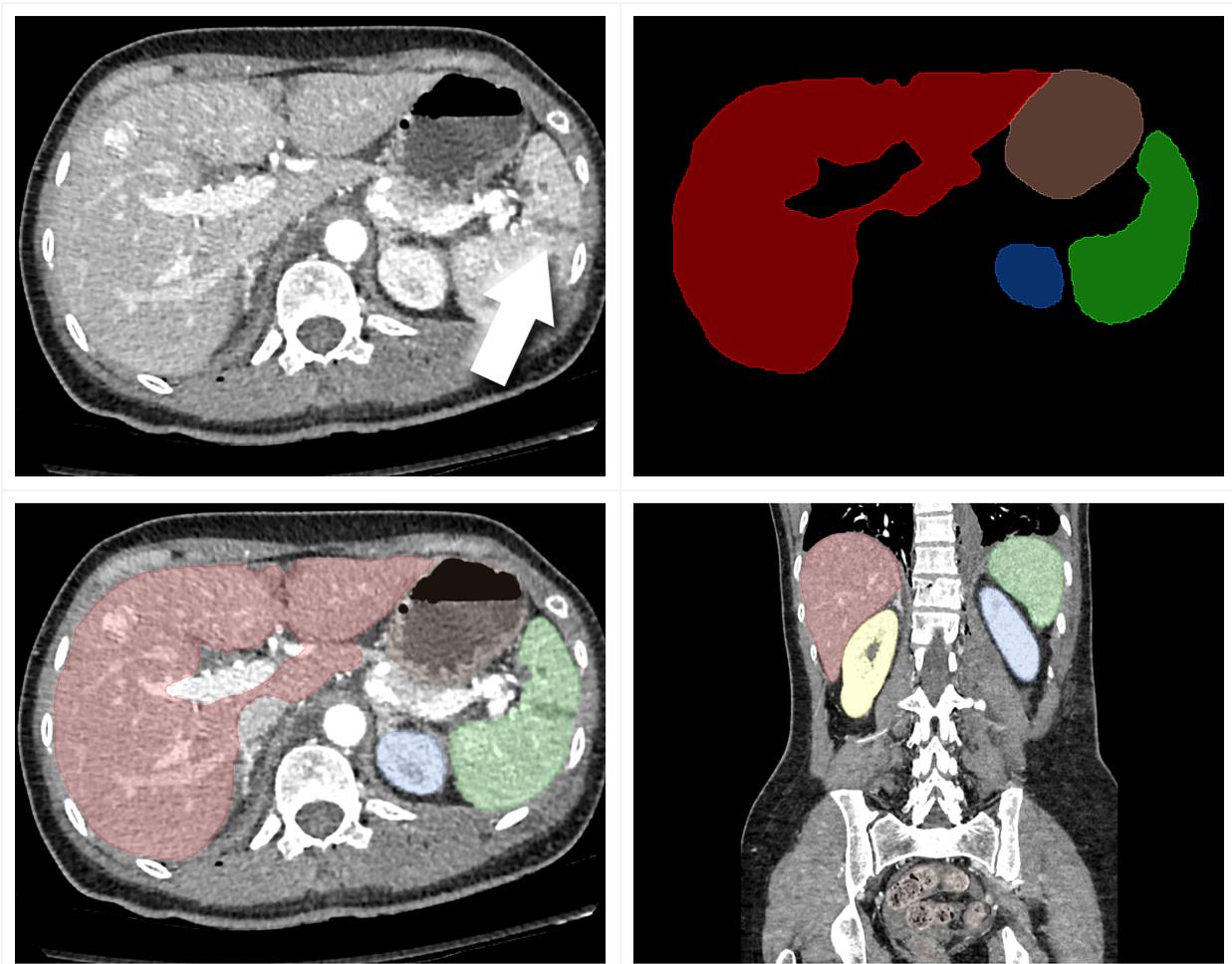

**Figure 2**. Example of abdominal organ segmentation with each color representing different organs. (A) Axial CT DICOM image demonstrating a splenic laceration (arrow). (B) Image illustrating the segmentations for the liver (red), spleen (green), left kidney (blue), and gastrointestinal tract (brown) in the axial plane. (C) Segmentation masks overlaying the corresponding CT image. (D) Segmentation masks overlaying the corresponding organs on a reconstructed coronal CT DICOM image. DICOM = Digital Imaging and Communications in Medicine.

## Supplementary Materials

**Data Collection**

An initial survey of potential contributing sites was conducted to determine the protocols used for the imaging of abdominal traumatic injuries. This survey demonstrated significant variation in imaging protocols used in terms of anatomical coverage, phases of imaging, and slice thickness. This survey also indicated that a strict inclusion criteria would result in a substantially smaller dataset when compared to broader inclusion. To help ensure a representative dataset that could lead to the development of more generalizable models, arterial, portal venous, split bolus, and multi-phasic protocol imaging were all deemed acceptable. Non-contrast imaging was not included as it was felt to offer little value to model training and is not a standard practice at most institutions. While delayed phase imaging is important in evaluating integrity of the collecting system, the majority of the potential contributing sites did not perform it routinely. Furthermore, the organizing committee expressed reservations about introducing bias and the potential for models to form spurious correlations if delayed phase imaging was more common in scans with renal injuries. The minimal acceptable anatomical coverage for these phases of imaging was the entire liver while no upper bound on anatomical coverage was defined. The maximum allowable slice thickness was 5.0 mm with a preference for thinner slice images. The minimum patient age was 18 years and post-laparotomy scans were excluded. No patient was to be represented more than once in the submitted scans. Details about compliance of local ethics regulations, scan identification, and de-identification were left to the discretion of each contributing site based on their own local privacy policies.

Contributing sites were asked to enrich the dataset with 50% of submitted cases with at least one of the following: solid organ injuries (liver, splenic, right renal, left renal), bowel and mesenteric injuries, or active extravasation. The remainder of the dataset was to be split equally between normal cases and cases with traumatic findings but none of the above listed injuries. The contributing sites were requested to submit de-identified CT images in DICOM format. These sites were requested to submit a comma separated values file which included patient sex and age, as well as binary labels indicating the presence or absence of solid organ injuries (liver, splenic, right renal, left renal), bowel and mesenteric injuries, and active extravasation with information being extracted from the clinical radiology report. A brief note describing the site(s) of active extravasation was requested to help focus subsequent expert annotation of the scans. Providing AAST grading of the solid organs was not necessary for the contributing sites.

**Standardization of Anatomic Coverage and Determination of Imaging Phase**

The contributed CT scans had extensive variability in anatomical coverage which ranged from imaging of the upper abdomen to continuous imaging from the skull vertex to toes. Manual cropping of the scans to a standardized range was deemed not feasible given the size of the dataset, available human resources, and time constraints. An automated pipeline (22) was used to crop the CT scans with an upper bound of the mid-heart and lower bound of the proximal femur. The deep learning-based algorithm was conditioned to map each voxel in a CT volume to its corresponding coordinate within a learned whole-body standard atlas space. In turn, these predictions are used to project fixed upper and lower field-of-view bounds, defined once in atlas space, onto each individual exam. The phase of imaging is not readily apparent based on series descriptions and time stamps of the submitted DICOM images. The same automated pipeline was utilized to determine aortic attenuation in Hounsfield units to serve as a surrogate for the phase of imaging. The processed scans were manually reviewed by a radiologist to ensure appropriate anatomic coverage. This review showed that scans which were imaged from the skull to feet often included more of the thorax than desired following processing but it still represented a dramatic improvement in coverage when compared to the original scan.

**Technical and Visual Validation**

A technical validation process was performed on each contributed study to ensure DICOM integrity and fidelity. This involves verifying that every DICOM file in a series has been assigned identical values for the specified DICOM elements, namely: Accession Number, Acquisition Number, Bits Allocated, Bits Stored, Columns, Frame of Reference UID, Patient ID, Pixel Spacing, Rows, Series Number, Slice Thickness, Study Date, Study Time, and Study Instance UID. In addition, any series with missing images or extra images using the Instance Number and Patient Image Position DICOM elements were flagged for review. Any DICOM images that were corrupted or not in the axial plane were removed. Each exam was scrutinized to ensure it had a distinct Patient ID, Accession Number, and Frame of Reference UID to mitigate the risk of data leakage.

Scans were manually reviewed by a radiologist as an opportunity to flag scans for removal from the dataset. Any CT series with insufficient coverage of the abdomen (i.e. less than the entire liver), incorrect phase (e.g. non-contrast, delayed phase), acute post-operative changes, non-soft tissue window images, and significant motion artifacts were removed from the dataset. An "incomplete FOV" was applied to CT series where the entire liver was imaged but not the entirety of the spleen and/or kidneys.

**Annotation of Cases**

Only CT scans flagged as positive by the contributing sites for the relevant abdominal traumatic injuries underwent annotation. Scans without these injuries did not undergo review by the annotation team as contributing sites had access to clinical reports, which was deemed as being superior to volunteer annotators in establishing ground truth for these scans.

A call for volunteer annotators was sent by email to the membership of the American Society of Emergency Radiology and the Society of Abdominal Radiology. Radiologists with subspeciality training and/or professional experience in abdominal trauma were invited to participate as annotators. Volunteers were provided with a written guide, instructional video, and if relevant, an AAST injury grading scale. A set of 12 training cases was provided to each volunteer and those that passed 9 of the training cases were able to continue as annotators. The ground truth of the training cases was established by members of the organizing committee. The training sample for solid organ injuries included both positive and negative cases with representation of each AAST grade. Training for active extravasation and bowel/mesenteric injuries included an equal number of positive and negative cases.

Each label was annotated independently by multiple radiologists as prior AI challenges showed that a single labeler can lead to poor quality annotations. A majority vote for each label established the ground truth.

Annotators were assigned to one type of annotation (e.g. liver, spleen, active extravasation) rather than having them annotate every label for each scan. The organizing committee felt that was more efficient for annotators as they did not require to constantly switch between AAST grading scales. A web-based annotation platform (md.ai, New York) which shares many features of a Picture Archiving and Communication System (PACS) workstation was used. Rather than assigning a fixed number of cases to annotators, annotations were performed in a crowd sourcing mode where volunteers were free to determine how many scans they wanted to annotate. A public scoreboard that ranked annotators was used to motivate the volunteers.

*Liver, Spleen, and Kidneys*

The exam level annotation of liver, spleen, and renal injuries provided by the contributing sites was not made available to annotators. Each solid organ injury underwent annotation independently by 3 radiologists using the AAST grading system. Annotators had the option to dispute the contributing site label for scans they deemed as being normal. The individual AAST gradings were collapsed into low (AAST I - III) and high grade (IV and V) injury groups due to poor inter-rater observation for assigning AAST scores (8) and to facilitate model training. In scenarios where all 3 annotators assigned different gradings (i.e. no injury, low grade, and high grade), a member of the organizing committee adjudicated the case and assigned the final ground truth label. For multi-phasic imaging, the solid organ injury labels were applied to every series.

*Bowel/Mesenteric Injury and Active Extravasation*

Annotators were instructed to provide image level labels for bowel/mesenteric injuries and active extravasation. For the active extravasation annotation task, 3 radiologists independently reviewed each scan. Bowel/mesenteric injuries underwent independent review by 2 radiologists rather than 3 because of limited resources and time constraints. The annotators were instructed to apply image level labels to both series in a multi-phasic scan. Any image that was labeled as positive by at least 2 annotators was considered as positive for active extravasation and by at least 1 annotator was considered positive for bowel/mesenteric injury.

**Segmentation**

To provide anatomical context to injuries, voxel-level segmentations of abdominal organs on a subset of the dataset are provided. These cases were enriched for the presence of high-grade or multiple injuries to better cover atypical appearances of injured organs. Cases with two or more injuries were selected for segmentation using stratified random sampling, where injuries included renal (left and/or right), liver, splenic, bowel/mesenteric, and active extravasation. Samples were stratified on the type of injury, site, age group (18-35, 36-55, 55+), and sex. For sampling purposes, cases missing age or sex were randomly assigned to an age group or sex, respectively. Of the cases with two or more injuries, we sampled cases with at least one severe (high grade) injury and then also sampled cases with only low grade injuries. In total, 148 studies and 245 series with two or more injuries were sampled for segmentation. Thirteen studies and a total of 30 series were removed from the sampled pool due to errors in converting them to NIfTI format. Nine series from seven studies were removed due to segmentations failing as a result of cases having DICOM PixelData processed with a different binary value representation. A total of 206 segmentations (128 studies and 206 series) are provided, comprising 92 cases with at least one severe (high grade) injury and 36 cases with only low grade injuries.

nnU-Net (18), a 3D fully-convolutional network (3D Res-U-Net), was used to train a preliminary network on the TotalSegmentator dataset (19), focusing only on the organs being evaluated in the challenge. This consisted of (1) liver, (2) spleen, (3) left kidney, (4) right kidney, and (5) bowel. Bowel represented a combination of esophagus, stomach, duodenum, small bowel, and colon. The 3D patch size was auto selected to be 128 x 128 x 128 (x, y, z dimensions) and training was performed on a RTX 3090 GPU (CUDA version 11.2; NVIDIA, Santa Clara, CA; 24 GB memory) for 1,000 epochs using a combination of cross entropy and Dice loss function (1:1). This network was applied to the 206 training cases and then manually corrected by one of five radiologists (JDR, EC, PL, PC, BM with 8, 14, 1,5 and 2 years of experience respectively).

**Data Structure**

CT images in DICOM format are organized with a [patient_id]/[series_id]/[image_instance_number].dcm hierarchy. Study level injury annotations and demographic information are provided in CSV files. Patient_id is a unique study level identifier and matches the last token of the Study Instance UID. For example, for a Study Instance UID value of 1.2.123.12345.1.2.3.201, the Patient ID would be 201. Similarly, series_id is a series level identifier that matches the last token of the Series Instance UID. Image_instance_number corresponds to the Instance Number data element and indicates the position of an image within the series.

The dataset contains 141 CT scans (95 in the training set, 9 in the public test set, and 37 in the private test set) where DICOM PixelData was processed with a different binary value representation that will appear unusual when opened using the pydicom Python library (27) or other DICOM viewers. The organizing committee decided against altering the DICOM PixelData to remediate this issue in order to maintain DICOM file integrity. Instead, a potential solution was provided on the Kaggle platform (https://www.kaggle.com/competitions/rsna-2023-abdominal-trauma-detection/discussion/427217).

The train_demographics_2024.csv file contains information about patient demographics. Patient_id is a unique study level identifier and matches the last token of the Study Instance UID. The Age and Sex columns correspond to patient age and sex. Patient ages above 89 have been replaced with 90+ to maintain compliance with HIPAA.

The train_series_meta_2024.csv file contains information regarding the phase of imaging and anatomical coverage of each CT series. The aortic_hu column indicates the attenuation value of the abdominal aorta in Hounsfield units which acts as a proxy for the phase of imaging. For a multiphasic CT scan, the higher value indicates arterial phase imaging and the lower portal venous phase imaging. The incomplete_organ column indicates if any of the liver, spleen, or kidneys were not completely imaged for a particular series.

The train_2024.csv file contains information about traumatic abdominal injuries for each scan. The liver, spleen, and kidney columns can contain one of three values: healthy (i.e. not injured), low grade injury, and high grade injury. The bowel and extravasation columns are binary labels that indicate bowel/mesenteric injury and active extravasation respectively. The any_injury column indicates if any of the liver, spleen, kidney, bowel/mesentery, and active extravasation are positive for injury.

The image_level_labels_2024.csv file provides image level labels for bowel/mesenteric injuries and active extravasation. The injury_name column can have one of two values, Active_Extravasation or Bowel, and indicates if that particular injury is present on the image with the corresponding Instance Number in DICOM metadata.

The segmentation files are named according to Series Instance UID and represent a subset of the training set. Segmentation labels have the following values: (1) liver, (2) spleen, (3) left kidney, (4) right kidney, and (5) bowel consisting of esophagus, stomach, duodenum, small bowel and large bowel. Please note, the NIFTI and DICOM files may not be in the same orientation. Header information in both NiFTI and DICOM metadata should be used to determine the appropriate orientation. Aligning the orientation of the NiFTI and DICOM file data requires consideration of the affine matrix.

The train_dicom_tags_2024.parquet file provides the metadata extracted from the header of every DICOM image in the dataset.

**Kaggle Competition**

Due to the heterogeneity of the dataset, where each institution exhibits distinct DICOM metadata features, data harmonization is crucial for standardizing data input. A white list approach is used to address this step by restricting the number of DICOM elements and acts to perform another step of de-identification. The selected DICOM elements to retain are: Bits Allocated, Bits Stored, Columns, Content Date, Content Time, Frame of Reference UID, High Bit, Image Orientation Patient, Image Position Patient, Instance Number, kVP, Patient ID, Patient Position, Photometric Interpretation, Pixel Data, Pixel Representation, Pixel Spacing, Rescale Intercept, Rescale Slope, Rows, SOP Instance UID, Samples Per Pixel, Series Instance UID, Series Number, Slice Thickness, Study Instance UID, Window Center, and Window Width. All identifying DICOM elements, including Patient ID, Study Instance UID, Series Instance UID, and SOP Instance UID were de-identified using the pydicom Python library (24). Additionally, given the substantial size of the dataset, each DICOM file underwent compression using the Run-Length Encoding Lossless (RLELossless) method.

The dataset was partitioned into train, public test, and private test sets and included 723 cases in the private test set, 404 cases in the public test set, and 3,147 cases in the training set. All segmentation cases were assigned to the training set. Similar to sampling for segmentation cases, test set cases were selected using stratified random sampling, stratified on injury for positive cases and for both positive and negative cases, stratified on site, age group (18-35, 36-55, 55+), and sex. For sampling purposes, injuries included kidney (left and/or right), liver, splenic, bowel/mesenteric, and active extravasation. Cases missing age or sex were randomly assigned to an age group or sex during the sampling process. A single site did not include any positive cases so all cases from that site were assigned to the training set. From cases not used in segmentation, we sampled cases with two or more injuries, one set with at least one high grade injury and another set with only low grade injuries. Next, negative cases and positive cases with only one injury were sampled using stratified random sampling as described above, with negative cases stratified only on site, age group, and sex. After these sets were assigned to the test set, an additional 500 negative cases not yet assigned to train or test sets were randomly sampled for inclusion in the test set. To ensure a reasonable number of training cases for bowel extravasation, some cases initially assigned to the test set were randomly selected and assigned to the training set. All other negative cases were assigned to the training set. From the test set, stratified random sampling was used to select approximately 65% of the cases for the private test set with the remaining cases assigned to the public test set.

**Study Identification, Data Extraction, and Data De-identification**

Contributing sites are listed in alphabetical order and do not correspond to the order in Table 1.

_Alfred Health, Australia_

The institutional archive was searched for all contrast-enhanced (split bolus) abdominal CT scans performed for trauma between January 2020 and January 2023. The URN, Age, Sex, Accession Number, Report Text and Interpretation Code were retrieved. These studies were filtered using the interpretation code (normal, abnormal, and critical) to assist with classification. Radiology reports were reviewed and findings abstracted as present or not present. When extravasation was present, the location, series and image number were recorded. Studies were sent to XNAT (NRG, Washington University School of Medicine) and de-identified using a DicomEdit script and a custom written Python script. The desired axial series were manually identified and exported from XNAT.

### Chiang Mai University, Thailand

The Faculty of Medicine at Chiang Mai University searched their PACS backup archive (Synapse Radiology PACS version 5.7.000; FUJIFILM Medical systems) using RIS (Envision.Net) for CTs of the abdomen performed on patients with abdominal trauma in the emergency room between January 1, 2018 and December 31, 2022. Radiology findings were reviewed by two radiologists. Images were downloaded from the PACS in DICOM format.

### China Medical University Hospital (CMUH), Taichung, Taiwan

This study was approved by the Ethics Committee and Institutional Review Board of CMUH (IRB No. CMUH112-REC1-074). The Big Data Center of CMUH obtained all data from the iHi Data Platform that includes the carefully verified electronic health data of administrative and demographic information; diagnoses, medical and surgical procedures, and prescriptions; laboratory measurements; physiologic monitoring records; and data on hospitalization; catastrophic illness status; registry data; and National Death Registry from patients who sought care at CMUH between 2003 and 2020.

Patients with trauma who received care in the Emergency Department (ED) of CMUH between 2012 and 2020 were initially included from the CMUH's Trauma Registry. To identify the source population of patients with organ injuries in the abdomen, we selected patients aged 18 years or older who had the ICD diagnosis codes for abdominal organ injuries (ICD-9: 863, 864, 865, 866, 867, 868) and underwent an abdominal computed tomography (CT) scan during their emergency visit. To identify the source population of adult patients with bone fractures, we excluded those who had organ injuries-related ICD codes during the same ED visit and kept those who had the ICD diagnosis for rib, lumbosacral spine, femur, and pelvis fractures (ICD-9: 805, 806, 807, 808, 820, 821). To obtain the source population of control patients without any organ injury or fractures, we excluded all patients admitted to ED with ICD codes for organ-related injuries or fractures and those without any CT scan during their ED visits. Next, for each source population, we manually reviewed the CT reports and annotated the key words of organ injuries (e.g., hepatic, splenic, renal, and bowel or mesenteric injuries, or active extravasation), traumatic injuries to other structures without organ injury, and negative for any traumatic injury. Furthermore, we kept only the records from patients who had consented on providing data for research use or who had died before the data retrieval date. Finally, we randomly selected cases with 50% being organ injuries cases, 25% fracture cases, and 25% normal controls. The DICOM images of the axial and soft tissue window kernel were downloaded from the institutional PACS.

The de-identification process complied with ISO-certified CMUH's data security specifications pipeline for ISO-29100, ISO-29191, and ISO-27001. We performed irreversible hash calculations on the patient's personal information in the DICOM file (including name, birthday, medical record number, national identity number, and any identifiable information), and re-established an index value based on the serial number to preserve links between CT data. A total of 42 DICOM tags were retained, all of which were imaging parameters during CT acquisition such as kVP.

_Clinica Santa Maria, Santiago, Chile_

This retrospective study received institutional approval with a waiver for informed consent. A database search was performed using the institutional RIS (AGFA Healthcare Enterprise Imaging; version 8.1.2). The search included abdomen and pelvis contrast-enhanced CTs obtained between January 1, 2017 and March 31, 2023, with the following keywords: fracture, laceration, contusion, hematoma, active bleeding, bleeding, or active extravasation. Radiology reports were manually reviewed to assess exclusion criteria, including no history of trauma, recent surgery, and patient age less than 18 years. Reports were classified as positive or negative for liver injury, splenic injury, renal injury, bowel or mesenteric injury, active extravasation, hemorrhage or traumatic fluid collections, lower rib fractures, spinal fractures, and proximal femora or pelvis fractures. CT scans with equivocal reports were excluded. Positive CT scans and a proportional number of negative scans were downloaded from the institutional PACS and de-identified. A body radiologist reviewed the extracted images to ensure optimal image quality and the absence of private health information.

*Eberhard Karls University Tübingen, Germany*

University Hospital Tuebingen searched their institutional RIS (mesalvo, Mannheim, Germany) for emergency department contrast CT scans performed between 2018 and 2023. Studies were included according to inclusion criteria and radiology reports were manually reviewed and categorized as positive or negative for trauma. CT scans with equivocal reports were excluded. Studies with axial section thickness greater than 3.0 mm were excluded. For all cases, patient sex and age were collected. For studies positive for trauma, additional information on injuries were extracted from the report. Images in the axial plane were downloaded in DICOM format and de-identified from PACS (syngo.via, Siemens Healthineers, Erlangen, Germany) to a central workstation.

*Gold Coast, Australia*

Following approval by the Gold Coast Hospital and Health Service, a PACS search to identify all trauma CT whole body scans of Emergency Department patients performed between January 1, 2013, and December 31, 2022. Studies not meeting inclusion criteria (age > 18 years, biphasic protocol covering solid abdominal organs) were discarded. The study reports were manually reviewed and scans were categorized as either positive or negative for visceral or bony injury, and further classified by type of injury. Thin slice axial sequences were downloaded from the PACS in DICOM format and underwent de-identification.

*Hospital Universitario Ramón y Cajal, Spain*

With the approval of the Hospital Ramón y Cajal review board, reports from emergency department abdominal and pelvis CT scans with intravenous contrast performed between January 2013 and March 2023 were exported from the institutional archive and manually reviewed by a board-certified radiologist and a senior resident. Reports were classified as positive or negative for traumatic abdominal injury. The specific abdominal traumatic injuries were recorded for each positive scan. Positive CT scans and an equal number of negative scans that include axial images (maximum of 3.0 mm) with arterial and venous phase were downloaded from PACS (Fujifilm Synapse 3D v6.1) and de-identified. Each CT scan was manually reviewed to assess diagnostic image quality and to ensure the absence of protected health information.

*Koç University, Türkiye*

Following approval of the Koç University institutional review board, a search was performed of the institutional PACS (Sectra IDS7; Sectra AB) for contrast-enhanced abdominal CT studies containing the word "trauma" obtained between 2016 and 2023. The corresponding radiology reports were manually reviewed by a radiologist and categorized as positive or negative for abdominal trauma findings. CT scans with equivocal reports and patients younger than 18 years of age were excluded. In all cases, data including patient age, study date, and patient sex were collected. For the positive cases, the locations of positive trauma findings were noted. Axial images were anonymized and manually exported from the PACS as DICOM images. The images were then uploaded to MD.ai for final curation and processing.

*Kingston Health Sciences Centre and Queen's University, Kingston, Canada*

Kingston Health Sciences Centre retrospective search was performed using the institutional RIS (GE Healthcare's Centricity Universal Viewer) and Nuance mPower (Nuance Communications) for contrast-enhanced CT of the chest, abdomen, and pelvis requested by the emergency department, and obtained between June 1, 2012, and April 25, 2023. Prior approval of the Queen's University Health Sciences & Affiliated Teaching Hospitals Research Ethics Board was obtained. The search keyword was "trauma." Images were reviewed by an abdominal radiologist, and cases which were not obtained in the setting of acute trauma or did not utilize the institutional trauma protocol were excluded. The location and grading of injuries was documented. Images of positive cases and negative cases were downloaded from the PACS (GE Healthcare's Centricity Universal Viewer) using RSNA Anonymizer. Custom scripts written in the Python programming language were used to extract the axial arterial and portal venous phase series with a section thickness of 2.5 mm or less. Images were de-identified using RSNA Anonymizer. Extracted images were manually reviewed to ensure the absence of protected health information.

### *Marrakech University Hospital, University Caddi Ayyad, Morocco*

Faculty of Medicine and Pharmacy of Marrakech, Caddi Ayyad University searched their PACS backup archive (Syngo Plaza, Siemens) using RIS (Hosix.net) for abdominal trauma CT obtained between January 1, 2021 and December 31, 2022. Radiology reports were searched for a specific keyword, manually reviewed, and categorized as positive or negative. Images were downloaded from the PACS in DICOM format. Images were de-identified using RSNA Anonymizer.

### *Mater Dei Hospital, Malta*

A retrospective search was performed using the institutional RIS (GE Healthcare's Centricity Universal Viewer) for CT poly-trauma (CTPOLTRA code) studies performed between January 2015 and February 2023, including patients that were 18 years of age or older. Radiology reports were individually reviewed and categorized as positive or negative for acute intra-abdominal injuries. The positive cases were then documented according to the types of injuries present. A spectrum of injury severity was also obtained. Positive and negative CT cases were downloaded and anonymized from the local PACS (GE Healthcare's Centricity Universal Viewer), which were then again de-identified using RSNA Anonymizer. Each of the extracted studies was reviewed in order to ascertain the absence of protected health information.

### *Medical College of Wisconsin*

A process for extracting, preparing, and analyzing computed tomography (CT) scans from emergency department (ED) patients for radiological review was developed. The initial step involved the execution of an SQL query to the electronic medical record (Epic Clarity) to retrieve data of subjects who underwent abdominal, pelvis, or chest CT scans with multi-volume axial scans. Following this, the corresponding DICOM images for each subject were accessed from the PACS system via a research-dedicated query/retrieve process. De-identification proceeded using Pandas dataframe manipulation in conjunction with Python's hash function. This process resulted in two distinct spreadsheets: one containing participant keying information and the other featuring only hashed IDs alongside corresponding reports, supplemented with additional columns for RSNA-specified injuries and biological landmarks. Radiologists then labeled the correct injuries and landmarks for each subject.

### *Mount Sinai Health System, New York City*

Mount Sinai Health System utilized Nuance mPower to search for all CT studies between January 1, 2016 and February 1, 2023 leveraging our department's dedicated trauma CT angiography exam description. The resultant studies were then annotated manually by reviewing reports for findings included in the dataset inclusion criteria. Patients younger than 18 and older than 89 years-old were excluded. Corresponding accession numbers with radiology reports facilitated batch transfer of image data to an on-premises XNATs server and de-identified by stripping all DICOM meta-data tags containing personal health information. A key was created using anonymized study IDs linked to radiology report annotations.

### NSW Health, Australia

A retrospective search was conducted within our institutions PACS for CT abdominal and pelvic trauma imaging. Each report was manually reviewed and categorized into positive or negative for trauma findings. These were then further divided into subgroups based on the location of the trauma (liver, spleen, renal, bowel, extravasation, peritoneal, rib, lumbosacral, pelvic, femoral). Inclusion criteria were > 18 years of age and CT images conducted using a biphasic protocol. Axial DICOM images were downloaded and anonymized using the RSNA Anonymizer tool.

### Tallaght University Hospital, Dublin, Ireland

This project received institutional approval with a waiver for informed consent. A data protection impact assessment was carried out and approved by our local data protection office. A retrospective search of our institutional PACS for CT of the thorax, abdomen and pelvis performed with the clinical indication of trauma performed between January 2020 and December 2022. CT scans without biphasic protocol assessing the upper abdominal solid viscera and scans of patients younger than 18 years of age were excluded. Reports were reviewed and relevant findings were recorded on a spreadsheet. A selection of normal and abnormal studies was made. Images were downloaded from PACS and de-identified using anonymizer software. A further round of de-identification was then performed on RSNA Anonymizer.

### Thomas Jefferson University Hospital, USA

Following approval of the Thomas Jefferson University institutional review board, a corpus of radiology reports between 2021 and 2023 was extracted from the dictation system (Nuance Powerscribe, Burlington, Vermont) using the mPower natural language processing search tool.

The searches were filtered for all emergency department encounters with an indication of "trauma" using two CT body examination codes (CT CHEST ABDOMEN PELVIS WO CONTRAST and CT CHEST ABDOMEN PELVIS WITH CONTRAST) over a two year for 15 hospitals in the Jefferson network, yielding a more heterogeneous population and set of acquisition devices. Each search query was expanded to be inclusive of a variety of injuries of each type (e.g. INDICATION:("trauma") & IMPRESSION:("liver injury" or "liver laceration" or "liver contusion" or "subcapsular hematoma" or "AAST" or "portal vein injury" or "hepatic arterial injury"). Eight searches were performed, each focusing on a specific feature needed for the dataset: liver, splenic, renal, bowel injuries, active extravasation, hemorrhage, and fracture mentioned in the impression of the report only. A final search for controls was performed where "no acute injury", "no acute abnormality" or "no traumatic abnormality" was found in the impression section of the report. Each search yielded between fifty and one-hundred exams. The exams were then randomized and combined, and duplicates were removed yielding one exam per patient. The reports were double-checked and coded for each of the eight search categories. The adjudicated examinations were then extracted from the PACS archive (Philips Intellispace) and de-identified using RSNA Anonymizer. The RSNA Anonymizer utility also removed any secondary capture objects from the examinations, as well as any scanned documents or dose images.

_Universidade Federal de São Paulo, Brazil_

All radiology reports from abdominal CT scans performed between 2012 and 2021 were exported from the RIS and searched for trauma-related words to pre-label the studies. Reports were manually reviewed to correct any inconsistencies in the pre-labeling process while labeling the organs affected by the trauma. The studies that met the inclusion criteria were downloaded from PACS using an in-house developed script.

*Unity Health Toronto, Canada*

Unity Health Toronto searched their institutional RIS (syngo, Siemens Medical Solutions USA) using Nuance mPower (Nuance Communications) for contrast enhanced trauma CT chest, abdomen and pelvis studies obtained between January 1, 2018 and April 15, 2023. The studies that met the inclusion criteria were manually reviewed by a radiologist and categorized as positive or negative for the relevant traumatic abdominopelvic injury. The location and grade of injury was recorded in positive cases and a balanced dataset was created. Patient age, sex, and study date were collected for all studies. Axial soft tissue window images were downloaded from PACS (Carestream PACS, Carestream Health, Rochester, NY, U.S.) in DICOM format and de-identified using RSNA Anonymizer.

*University Hospital Würzburg, Würzburg, Germany*

With approval from the local institutional review board (IRB number: 20230227 01), two radiologists of University Hospital Würzburg with 6 and 9 years of training in the field searched the institutional database for biphasic trauma scans (late arterial phase covering the thorax and solid abdominal organs plus portal venous phase of the abdomen and pelvis). Radiology reports between September 2019 and December 2022 were manually reviewed. The presence of liver, splenic, renal, bowel, lower rib, lumbosacral spine, pelvic, and proximal femur injuries were determined in dichotomous fashion (present/absent). In addition, the presence and location of active extravasation were noted as well as the presence of hemorrhage/traumatic fluid. Patient age and sex were also recorded. CT scans with equivocal reports, a different scan protocol, or of patients younger than 18 years were excluded. A random sample of negative cases that equaled the number of positive scans was added to generate a balanced dataset. Images in axial orientation were downloaded from the local PACS (Merlin, Phönix-PACS) in DICOM format and de-identified using RSNA Anonymizer.

*University Hospitals Cleveland Medical Center, USA*

Following approval of the University Hospitals Cleveland Medical Center institutional review board, our research team queried our PACS (Sectra) for exams with the Current Procedural Terminology (CPT) code for CTs including the abdomen and pelvis with a trauma designation presenting to the emergency room between April 1, 2013 and April 1, 2023 for patients between the ages of 18 to 89. Access to these results were via the Power BI Cloud Service. The final radiology reports were consecutively reviewed and categorized as positive or negative for abdominal trauma as well as the specific location of the trauma. Equivocal findings were reviewed individually by returning to the original images by our research team. Additional data including the patient age, study date, and accession number were also collected. Images formatted in the axial, coronal, and sagittal planes were downloaded from the PACS to a central workstation where the studies were anonymized using the RSNA anonymizer tool.

*University of Sarajevo, Sarajevo, Bosnia and Herzegovina*

A comprehensive retrospective search was conducted using the institutional PACS, Radiology Information System, and Hospital Information System scans with contrast media acquired between January 2012 and March 2023. Rigorous peer-reviewed analysis of CT reports was undertaken to apply exclusion criteria, which included patients under 18 years of age and unenhanced CT scans. This meticulous process involved assessment by two experienced radiologists. For each positive case identified, meticulous documentation of the precise location of the injury was carried out.

*University of Utah, USA*

University of Utah searched their institutional RIS (Radiant; Epic Systems Corporation) using Nuance mPower (Nuance Communications) to identify CT abdomen and pelvis studies for the AI challenge. The searches used included the following: AAST, trauma AND "bowel injury",

"shattered kidney" OR "renal laceration" OR "kidney laceration", "shattered spleen" OR "spleen laceration" OR "splenic laceration". Since liver injury was relatively common among our patient population, there was not a dedicated search for liver injury. The results of this search were manually reviewed and categorized as either abdominal organ trauma, traumatic findings without abdominal organ injury, and normal CT scan in a trauma patient. Studies were dated from July 1, 2013 through May 18, 2023. Additional data including patient age (if 89 years old or younger) and patient sex were also collected. CT scans with equivocal reports were excluded. Images from the CT abdomen/pelvis exam in the portal venous phase formatted in the axial plane using the soft tissue kernel were downloaded from the PACS (IntelliSpace PACS; Philips Healthcare) as well as arterial phase CT abdomen/pelvis, arterial phase CT chest, or delayed phase CT abdomen/pelvis if available. Images were downloaded to a local workstation where the studies were anonymized.

### Vancouver General Hospital

Abdominal trauma cases from Vancouver General Hospital were collected in a systematic manner. The institutional Radiology Information System (RIS) at Vancouver General Hospital, was queried for multiphasic abdominal trauma CT scans performed between January 1, 2015, and December 31, 2022. The search was facilitated by the hospital RIS system with a focus on emergency department cases.

Subsequently, the collected data underwent a comprehensive review process. An emergency and trauma radiologist meticulously examined the radiology reports, categorizing them as either positive or negative for abdominal trauma. For positive cases, details regarding the patient age, gender and site of abdominal trauma were systematically recorded. Exclusion criteria were applied to ensure the integrity and relevance of the dataset. CT scans with equivocal reports, descriptions of prior abdominal surgeries were excluded from the study. To achieve a balanced dataset, a random sample of negative cases was selected, matching the number of positive scans. The actual CT images were retrieved in DICOM format from the PACS at Vancouver General Hospital. Prior to further analysis, all images underwent a rigorous de-identification process. The RSNA Anonymizer tool was employed to strip any personally identifiable information. To ensure compliance with privacy regulations and data security, a radiologist conducted a manual review of the de-identified images, confirming the absence of private health information, details of prior abdominal surgeries, and any indication of intravenous contrast media administration.